\documentclass[twoside,11pt]{article}

\usepackage{blindtext}

\usepackage[abbrvbib, preprint]{jmlr2e}

\usepackage{lastpage}
\jmlrheading{23}{2025}{1-\pageref{LastPage}}{1/21; Revised 5/22}{9/22}{21-0000}{Vineet Punyamoorty, Aditya Malusare and Vaneet Aggarwal}

\ShortHeadings{}{}
\firstpageno{1}

\usepackage{amsmath}
\usepackage{dsfont}
\usepackage{booktabs}
\usepackage{hyperref}
\usepackage{algorithm}
\usepackage{algpseudocode} 
\hypersetup{
    colorlinks=true,
    linkcolor=blue,
    filecolor=magenta,  
    citecolor=blue,
    urlcolor=blue,
    pdftitle={Contrastive Cross-Modal Learning for Infusing Chest X-ray Knowledge into ECGs},
    }
\newcommand{\modelname}{\textsc{CroMoTEX}\xspace}

\begin{document}

\title{Contrastive Cross-Modal Learning for Infusing Chest X-ray Knowledge into ECGs}

\author{\name Vineet Punyamoorty \email vpunyamo@purdue.edu \\
       \addr Elmore Family School of Electrical and Computer Engineering\\
       Purdue University\\
       West Lafayette, IN 47907, USA
       \AND
       \name Aditya Malusare \email malusare@purdue.edu \\
       \addr Edwardson School of Industrial Engineering\\
       Purdue University\\
       West Lafayette, IN 47907, USA
       \AND
       \name Vaneet Aggarwal \email vaneet@purdue.edu \\
       \addr Edwardson School of Industrial Engineering\\
       Purdue University\\
       West Lafayette, IN 47907, USA}

\editor{My editor}

\maketitle

\begin{abstract}
Modern diagnostic workflows are increasingly multimodal, integrating diverse data sources such as medical images, structured records, and physiological time series. Among these, electrocardiograms (ECGs) and chest X-rays (CXRs) are two of the most widely used modalities for cardiac assessment. While CXRs provide rich diagnostic information, ECGs are more accessible and can support scalable early warning systems. In this work, we propose \modelname, a novel contrastive learning-based framework that leverages chest X-rays during training to learn clinically informative ECG representations for multiple cardiac-related pathologies: cardiomegaly, pleural effusion, and edema. Our method aligns ECG and CXR representations using a novel supervised cross-modal contrastive objective with adaptive hard negative weighting, enabling robust and task-relevant feature learning. At test time, \modelname relies solely on ECG input, allowing scalable deployment in real-world settings where CXRs may be unavailable. Evaluated on the large-scale MIMIC-IV-ECG and MIMIC-CXR datasets, \modelname outperforms baselines across all three pathologies, achieving up to 78.31 AUROC on edema. Our code is available at \href{https://github.com/vineetpmoorty/CroMoTEX}{github.com/vineetpmoorty/cromotex}.

\end{abstract}


\section{Introduction}
\begin{figure*}
    \centering
    \includegraphics[width=\linewidth]{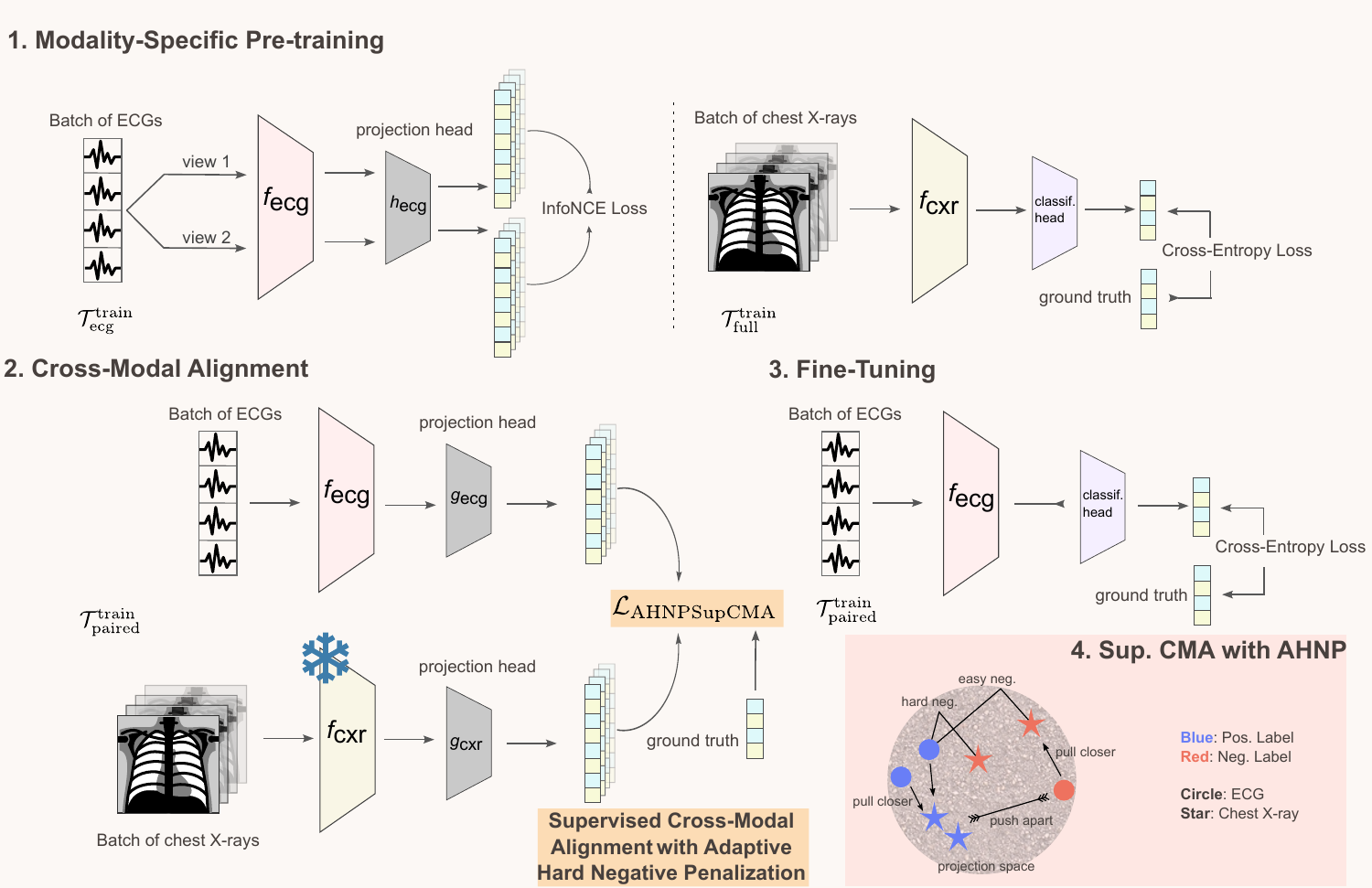}
    \caption{\modelname consists of modality-specific encoders $f_\text{ecg}$ and $f_\text{cxr}$, along with respective projection and classification heads. \textbf{1.} We first pre-train $f_\text{ecg}$ in a self-supervised fashion, and pre-train $f_\text{cxr}$ on the disease label. \textbf{2.} We next perform supervised cross-modal alignment by infusing the knowledge from $f_\text{cxr}$ to learn rich, task-relevant ECG representation and then \textbf{3.} fine-tune the ECG encoder on the disease labels. \textbf{4.} We propose $\mathcal{L}_\text{AHNPSupCMA}$, a novel loss function that uses Adaptive Hard Negative Penalization (AHNP) to allocate higher weights to the losses arising from hard negatives (Sec.~\ref{subsec:ahnp_text}).}
    \label{fig:main_fig}
\end{figure*}

Cardiovascular diseases are the leading cause of mortality worldwide, emphasizing the need for reliable and scalable diagnostic tools. The growing availability of multimodal medical data presents an opportunity for cross-modal knowledge transfer, which can enable earlier and more accessible diagnosis by leveraging inexpensive, widely available modalities in place of less accessible gold-standard tests. Among commonly used diagnostic modalities, chest X-rays (CXRs) provide detailed anatomical insights but require hospital-based imaging and specialized equipment, making them less accessible. Electrocardiograms (ECGs), in contrast, are widely available, especially with the rise of wearable technology, and facilitate early screening. However, ECGs alone often lack the anatomical detail necessary for robust diagnosis, limiting their effectiveness. Given the complementary nature of CXRs and ECGs, transferring knowledge from CXRs to ECGs could enhance ECG-based disease detection, improving early diagnosis and patient outcomes. Despite this potential, little research has explored this direction, with most prior research on multimodal ECG learning focusing on multimodal fusion with other modalities \citep{hayat_medfuse_2023,yao_drfuse_2024,thapa_more_2024} and alignment with clinical text for zero-shot classification \citep{liu_zeroshot_2024}. ECCL \citep{ding_crossmodality_2024} and Cross-modal Autoencoder \citep{radhakrishnan_crossmodal_2023} explore cross-modal transfer from Cardiac Magnetic Resonance Imaging (CMR) to ECG. However, they rely on CMR, a significantly more resource-intensive imaging modality compared to CXR, limiting its scalability and accessibility. Furthermore, they employ a simple alignment strategy that only aligns ECG-CMR pairs from the same sample, disregarding other semantically similar pairs that share the same disease label. Introducing supervision in the alignment process could help the model fully leveraging large labeled datasets, promoting it to learn more generalizable and task-relevant ECG representations. 

In this paper, we propose \modelname (\underline{\textbf{Cro}}ss-\underline{\textbf{Mo}}dal \underline{\textbf{T}}ransfer between \underline{\textbf{E}}CG and Chest \underline{\textbf{X}}-ray), a novel supervised cross-modal learning framework that facilitates knowledge transfer from CXRs to ECGs. To address the limitations of prior work, we propose a supervised contrastive alignment strategy that explicitly incorporates disease labels to guide ECG representation learning. To further improve the quality of alignment, we introduce a novel contrastive loss function with adaptive hard negative penalization (AHNP), which dynamically re-weights negative ECG-CXR pairs based on similarity, effectively penalizing harder negatives more and improving the alignment process. 

We evaluate our approach on three clinically significant conditions: {Cardiomegaly}, {Pleural Effusion}, and {Edema}. Cardiomegaly, or heart enlargement, is typically diagnosed via chest X-ray but may manifest subtle cardiac signals detectable in ECG. Pleural Effusion, the accumulation of fluid in the pleural cavity, and Edema, fluid retention often associated with heart failure, are both visually prominent on CXRs but may induce indirect electrical patterns measurable by ECG. Studying these pathologies allows us to investigate the viability of ECG-based prediction for conditions where CXRs provide diagnostic gold standards—making them ideal candidates for evaluating cross-modal knowledge transfer.

The key contributions of our work are:
\begin{enumerate}
    \item A novel supervised cross-modal learning framework for transferring knowledge from CXRs to ECGs, leveraging disease labels for task-relevant and generalizable representation learning
    \item A new contrastive loss function, $\mathcal{L}_\text{AHNPSupCMA}$, which adaptively penalizes harder negatives, improving ECG-CXR alignment
    \item Ablation studies validating the effectiveness of self-supervised ECG pre-training, supervised cross-modal alignment, and similarity-based hard negative penalization in our pipeline.
\end{enumerate}

\section{Related Work}
\subsection{Contrastive Learning}
Contrastive learning involves self-supervised learning of representation by maximizing the alignment between projected embeddings of augmented views of the same sample, while simultaneously minimizing the alignment between those of different samples, using the InfoNCE loss \citep{oord_2018_representation}. This enables the encoder to learn meaningful representations from datasets with little to no labeled data while promoting feature learning that generalizes effectively across diverse downstream tasks. Notable approaches include SimCLR \citep{chen_simple_2020}, which maximizes the agreement between augmented views using a simple framework with a large batch size and a projection head; MoCo \citep{he_2020_momentum}, which introduces a momentum-based memory bank to maintain a dynamic queue of negative samples for contrastive learning; and BYOL \citep{grill_2020_bootstrap}, which departs from explicit negative samples by relying on a momentum encoder and a predictor network to learn robust representations in a self-distilled manner. 

\subsection{Multimodal Contrastive Learning with Images}
Multimodal representation learning aligns samples from one modality with semantically related samples from another, enabling both to be represented in a shared embedding space. This facilitates downstream tasks such as cross-modal retrieval and zero-shot classification. A prominent example is CLIP \citep{radford_learning_2021}, which learns joint image-text representations by aligning images with their corresponding textual descriptions to enable flexible retrieval and zero-shot transfer. ALIGN \citep{jia_2021_scaling} extends this idea to massive image-text pairs, further improving generalization across diverse datasets. 

\subsection{Multimodal Learning in Healthcare}

Multimodal learning has emerged as a powerful paradigm in healthcare, enabling models to integrate heterogeneous data sources such as imaging, waveforms, structured clinical data, and free-text notes. The central goal is to improve clinical decision-making by leveraging the complementary strengths of different modalities.

GLoRIA \citep{huang_gloria_2021} is a foundational effort in this space, aligning chest X-ray images with their corresponding radiology reports through a contrastive learning framework. By jointly training vision and language encoders, the model learns clinically meaningful representations that enable zero-shot classification and improved report retrieval. This highlights how multimodal learning can bridge the gap between visual evidence and textual reasoning. Hager et al. \citep{hager_best_2023} extend this idea by fusing imaging data with structured clinical variables (e.g., age, sex, lab tests), demonstrating that integrating tabular features can enhance performance in disease classification tasks compared to image-only models. King et al. \citep{king_multimodal_2023} take a different approach by aligning continuous clinical measurements (e.g., vital signs, labs) with unstructured clinical notes using a contrastive loss. This strategy improves the quality of learned representations for ICU outcome prediction, even in the absence of labels, reinforcing the power of contrastive learning to extract useful supervision from weakly paired data.

Other recent studies such as Thapa et al. \citep{thapa_more_2024} and Ding et al. \citep{ding_crossmodality_2024} explore modality alignment at a fine-grained level. These works demonstrate the effectiveness of contrastive learning in aligning representations across diverse inputs such as EHR, ECG, and imaging. Together, these efforts suggest that multimodal learning—particularly when paired with contrastive objectives—holds significant promise for developing holistic, generalizable patient models.

\subsection{Contrastive Learning on ECGs and Chest X-rays}

Contrastive learning has proven to be a highly effective method for self-supervised representation learning, especially in medical domains where labeled data is scarce but raw signals and images are abundant. Within this context, ECGs and chest X-rays are two high-value modalities due to their ubiquity and diagnostic relevance.

For ECGs, models like CLOCS \citep{kiyasseh_2021_clocs} leverage contrastive learning on augmented views of raw ECG waveforms. By training on tasks such as segment-level and lead-wise discrimination, CLOCS learns representations that generalize well to downstream tasks including cardiovascular disease classification and arrhythmia detection. Similarly, Cheng et al. \citep{cheng_2020_subject} propose a subject-aware contrastive framework that improves ECG representations by contrasting signals from the same subject across time, which leads to better patient-level phenotyping.

On the imaging side, Sowrirajan et al. \citep{sowrirajan_moco_2021} propose MoCo-CXR, an adaptation of Momentum Contrast (MoCo) \citep{he_2020_momentum} for chest X-ray representation learning. Their model uses large-scale unlabeled CXR data to pre-train an encoder that significantly boosts performance on downstream classification tasks, such as identifying pulmonary conditions from frontal X-rays.

Additionally, CIA-Net \citep{liu_2019_align} introduces contrastive induced attention mechanisms to improve localization of abnormalities in CXRs. This model explicitly encourages attention maps to align across augmented views, improving both classification and interpretability.

These advances collectively illustrate how contrastive learning—whether applied to time-series data like ECGs or spatial modalities like chest X-rays—can uncover robust and transferable representations. In particular, the potential to align these two modalities under a unified contrastive framework opens new opportunities for cross-modal learning and generalization across diagnostic tools.

\section{Methods}
\subsection{Problem Setup: Supervised Cross-Modal Contrastive Learning}
We propose a combination of CLIP-style \citep{radford_learning_2021} cross-modal alignment and SupCon-style \citep{khosla_supervised_2020} supervised objective to perform supervised cross-modal contrastive learning as described below. Given a labeled dataset $\mathcal{D}_{\text{paired}}$ of ECGs ($e_i$), chest X-rays ($x_i$) and disease labels ($y_i$), $\mathcal{D}_{\text{paired}} = \left\{e_i, x_i, y_i\right\}_{i=1}^N$, our goal is to learn representations for the two modalities by contrasting positive pairs against negative pairs. As seen in Fig.~\ref{fig:main_fig}, we have modality-specific encoders $f_\text{ecg}$, $f_\text{cxr}$ and projection heads $g_\text{ecg}$, $g_\text{cxr}$ which project the embeddings onto a common space. Given a batch $\mathcal{B}$ of samples $\left\{e_i, x_i, y_i\right\}_{i=1}^B$, after obtaining the projected embeddings $z^e_i = g_\text{ecg}\left(f_\text{ecg}\left(e_i\right)\right)$ and $z^x_i = g_\text{cxr}\left(f_\text{cxr}\left(x_i\right)\right)$ one can train the network consisting of $f_\text{cxr}, f_\text{ecg}, g_\text{cxr}$ and $g_\text{ecg}$ using the following cross-modal alignment objective: 
\begin{equation}
    \mathcal{L}_{\text{CMA}} = \frac{-1}{2}\sum_{i=1}^B{\log{\frac{\exp{\left(z^e_i\cdot z^x_i \right / \tau)}}{\sum_{a \in A(i)}\exp{\left(z^e_i\cdot z^x_a / \tau\right)}}}}
    - \frac{1}{2}\sum_{i=1}^B\log{\frac{\exp{\left(z^x_i\cdot z^e_i \right / \tau)}}{\sum_{a \in A(i)}\exp{\left(z^x_i\cdot z^e_a / \tau\right)}}}
\end{equation}
where $A(i) = \mathcal{B} \setminus \left\{i\right\}$. This objective is a symmetrized version \citep{ding_crossmodality_2024} of InfoNCE loss \citep{oord_2018_representation} and it aims to align the projected embeddings of the two modalities from the same sample $(e_i, x_i)$ while simultaneously pushing apart the projections from different samples $(e_i, x_j)$ for all $i \ne j$. However, this objective does not use supervision from the available labels $y_i$ and therefore does not account for the task-relevant, semantically positive pairs from different samples: $\left\{\left(e_i, x_j\right): y_i = y_j, i \ne j\right\}$. To address this, following SupCon \citep{khosla_supervised_2020}, we have the objective $\mathcal{L}_{\text{SupCMA}} = \frac{1}{2}\mathcal{L}_{\text{SupCMA}}^{e, x} + \frac{1}{2}\mathcal{L}_{\text{SupCMA}}^{x, e}$ where
\begin{equation}
    \mathcal{L}_{\text{SupCMA}}^{m, n} = -\sum_{i=1}^B\frac{1}{|P(i)|}\sum_{p \in P(i)}\log{\frac{\exp{\left(z^m_i \cdot z^n_p / \tau\right)}}{\sum_{a \in A(i)}\exp{\left(z^m_i \cdot z^n_a /\tau\right)}}}
    \label{eqn:supcma_term}
\end{equation}
$P(i) = \left\{p \in A(i): y_p = y_i\right\}$ represents the set of all positive pairs obtained by matching the label $y$.

\subsection{Supervised Contrastive Loss with Adaptive Hard Negative Penalization}
\label{subsec:ahnp_text}
In the supervised objective described above (Eqn.~\ref{eqn:supcma_term}), a sample pair $(e_i, x_j)$ with different labels $(y_i \ne y_j)$ can be termed as a \textit{negative}. However, not all negatives are equal---\textit{hard negatives}, i.e. pairs $\left\{(e_i, x_j): e_i, x_j \in \mathcal{B}\right\}$ with high similarity $z^e_i \cdot z^x_j$ in the projection space while having different labels $y_i \ne y_j$, are particularly challenging. Therefore, it could help to prioritize the losses arising from hard negatives more strongly than those from easy ones. To address this problem, we propose to modify $\mathcal{L}_{\text{SupCMA}}$ by performing weighted summation of the negative-pair terms in the denominator of Eqn.~\ref{eqn:supcma_term}, such that hard negatives are assigned higher weights, which we call \underline{\textbf{a}}daptive \underline{\textbf{h}}ard \underline{\textbf{n}}egative \underline{\textbf{p}}enalization (AHNP): 
\begin{equation}
    \mathcal{L}_{\text{AHNPSupCMA}}^{m,n} = -\sum_{i=1}^B\frac{1}{|P(i)|}\sum_{p \in P(i)}\log{\frac{\left(1+\beta\mathds{1}_{i=p}\right)\exp{\left(z^m_i \cdot z^n_p / \tau\right)}}{\sum_{a \in A(i)}w_{i, a}\exp{\left(z^m_i \cdot z^n_a /\tau\right)}}}
    \label{eqn:ahnpsupcma_term}
\end{equation}
where $w_{i, a} = \mathop{\mathrm{w}}\left(z^m_i \cdot z^n_a /\tau\right)$ is the weighting function.

To implement adaptive hard negative penalization, we define a weighting function \( w_{i, a} \) that determines how strongly each negative sample \( a \in A(i) \) contributes to the denominator of the contrastive loss for a given anchor \( z^m_i \). We explore three different weighting strategies—\texttt{linear}, \texttt{topk}, and \texttt{exp}—each designed to emphasize harder negatives (i.e., those with higher similarity to the anchor) more than easier ones.
\begin{itemize}
    \item \textbf{\texttt{linear}} strategy: The weights are scaled linearly based on similarity. For each anchor \( z^m_i \), we compute the similarity with every other sample in the batch and linearly assign weights \( w_{i,a} \) such that samples with higher similarity receive larger weights. Specifically, the weights are distributed from 1 up to a maximum value \( \alpha > 1 \), with \( \alpha \) controlling the strength of this emphasis. 
    \item \textbf{\texttt{topk}} strategy: For each anchor \( z^m_i \), we identify the top \( k\% \) of samples (in terms of similarity) and assign them a fixed higher weight \( \alpha > 1\), while all other samples receive a default weight of 1. This sharply focuses the contrastive loss on the hardest negatives in the batch, leaving easier negatives largely unaffected. 
    \item \textbf{\texttt{exp}} strategy: We employ an exponential function to assign weights. Each weight is computed as \( w_{i,a} = 1 + \exp(\alpha \cdot z^m_i \cdot z^n_a) \), where \( \alpha \) controls the steepness of the exponential curve. This strategy creates a highly nonlinear weighting profile in which even small increases in similarity between anchor and negative can result in a significant increase in penalty.
\end{itemize}

In all three strategies, \( \alpha \) (and \( k \) for \texttt{topk}) are treated as tunable hyperparameters that govern how aggressively the model prioritizes hard negatives during training.

In addition to AHNP, we also assign higher weight $\beta$ ($\beta > 1$) to within-sample cross-modal positive pairs in the numerator of our contrastive loss, to encourage stronger alignment between inherently corresponding ECG and CXR representations. This extra weighting for within-sample cross-modal pairs allows us to tune the loss function between two extremes: one where no supervision from labels is used (at high values of $\beta$, the extra weight given to within-sample pairs) and another where all same-label pairs are treated equally.

We use the objective $\mathcal{L}_{\text{AHNPSupCMA}} = \frac{1}{2}\mathcal{L}_{\text{AHNPSupCMA}}^{e, x} + \frac{1}{2}\mathcal{L}_{\text{AHNPSupCMA}}^{x, e}$ in this work.

\begin{algorithm}[H]
\caption{Supervised Cross-Modal Contrastive Learning with Adaptive Hard Negative Penalization}
\label{alg:training}
\begin{algorithmic}[1]
\Require Paired dataset $\mathcal{D}_\text{paired} = \{(e_i, x_i, y_i)\}_{i=1}^N$, ECG-only dataset $\mathcal{T}_\text{ecg}^\text{train}$, pre-trained weights for $f_\text{cxr}$
\Require Hyperparameters: temperature $\tau$, weights $\alpha$, $\beta$, batch size $B$, top-$k$, learning rates

\Procedure{PreTrain}{$f_\text{cxr}, f_\text{ecg}$}
    \State Train $f_\text{cxr}$ on cardiomegaly labels using cross-entropy loss
    \State Train $f_\text{ecg}$ with SimCLR-style contrastive learning on $\mathcal{T}_\text{ecg}^\text{train}$
    \State Discard projection head $h_\text{ecg}$ after training
    \State Freeze $f_\text{cxr}$
\EndProcedure

\Procedure{CrossModalAlignment}{}
    \For{each training step}
        \State Sample batch $\mathcal{B} = \{(e_i, x_i, y_i)\}_{i=1}^B$ using weighted random sampler
        \State Compute $z^e_i = g_\text{ecg}(f_\text{ecg}(e_i))$, $z^x_i = g_\text{cxr}(f_\text{cxr}(x_i))$ for all $i$
        \For{each anchor $i = 1$ to $B$}
            \State $P(i) \gets \{p \neq i \mid y_p = y_i\}$
            \State $A(i) \gets \mathcal{B} \setminus \{i\}$
            \For{each $a \in A(i)$}
                \State Compute similarity $s_{i,a} \gets z^m_i \cdot z^n_a$
                \State Compute weight $w_{i,a}$ using chosen AHNP strategy:
                \If{\texttt{linear}} $w_{i,a} \propto$ rank of $s_{i,a}$
                \ElsIf{\texttt{topk}} $w_{i,a} \gets \alpha$ if $a \in$ top-$k$ similar; else $1$
                \ElsIf{\texttt{exp}} $w_{i,a} \gets 1 + \exp(\alpha \cdot s_{i,a})$
                \EndIf
            \EndFor
            \State Compute loss $\mathcal{L}_i$ using weighted SupCon-style formulation with $\beta$ emphasis on within-sample pairs
        \EndFor
        \State Compute total loss: $\mathcal{L}_\text{AHNPSupCMA} \gets \frac{1}{2} \sum_i \mathcal{L}_i^{e,x} + \frac{1}{2} \sum_i \mathcal{L}_i^{x,e}$
        \State Update $f_\text{ecg}$, $g_\text{ecg}$, $g_\text{cxr}$ using \texttt{AdamW}, cosine LR decay, and gradient clipping
    \EndFor
\EndProcedure

\Procedure{FineTune}{$f_\text{ecg}$}
    \State Discard projection heads $g_\text{ecg}, g_\text{cxr}$
    \State Fine-tune $f_\text{ecg}$ on cardiomegaly labels using cross-entropy loss
    \State Use same class-balanced sampling as in alignment phase
\EndProcedure
\end{algorithmic}
\end{algorithm}

\subsection{Model and Training Details}
\label{subsec:model_and_training_details}
Our model consists of an ECG encoder $f_\text{ecg}$, chest X-ray encoder $f_\text{cxr}$ and linear projection heads $g_\text{ecg}$ and $g_\text{cxr}$ (Fig.~\ref{fig:main_fig}). We implement $f_\text{ecg}$ as a series of two 1-D convolution layers, each consisting of ReLU activation and 1D \texttt{BatchNorm}, followed by a Transformer encoder consisting of 8 attention heads, 4 Transformer blocks \citep{vaswani_2017_attention} and 1D Adaptive Pooling to obtain the ECG embeddings $f_\text{ecg}(e_i)$. 

For the chest X-ray image encoder $f_\text{cxr}$, we use a DenseNet-121 model pre-trained on MIMIC-CXR data from TorchXRayVision \citep{cohen_2022_torchxrayvision}. We replace the original multi-label classification head with a single-label and further train it for disease label prediction.

\subsubsection{Pre-training}
\label{subsec:pretraining}
We first pre-train the modality-specific encoders $f_\text{ecg}$ and $f_\text{cxr}$. $f_\text{cxr}$ is pre-trained using $\mathcal{T}_\text{full}^\text{train}$ and $\mathcal{T}_\text{full}^\text{val}$ with Cross-Entropy loss on the disease labels using initial weights from TorchXRayVision \citep{cohen_2022_torchxrayvision}. We use the \texttt{Adam} optimizer with peak learning rate (LR) of $10^{-4}$, batch size 64, a learning rate scheduler with linear warm-up for 10 epochs from LR of $10^{-5}$, and gradual cosine decay back to $10^{-5}$. The backbone's LR is scaled by $0.05$ times relative to the classification head to preserve features learned from TorchXRayVision's pre-training. 

To pre-train the ECG encoder $f_\text{ecg}$, we perform self-supervised learning (SSL) using the datasets $\mathcal{T}_\text{ecg}^\text{train}$ and $\mathcal{T}_\text{ecg}^\text{val}$ following the SimCLR technique \citep{chen_simple_2020} to help it learn generalizable ECG features. We use a linear projection head $h_\text{ecg}$ to align the projected embeddings of two augmented views of each ECG in a batch using InfoNCE loss, following which $h_\text{ecg}$ is discarded. We use VCG-based augmentations for the pre-training, as described in Sec.~\ref{subsec:datasets}. 

\subsubsection{Cross-modal Alignment}
Once both the modality-specific encoders $f_\text{ecg}$ and $f_\text{cxr}$ are pre-trained, we freeze the chest X-ray encoder $f_\text{cxr}$ for all further steps. This ensures that $f_\text{cxr}$ serves as a teacher in the cross-modal alignment process, where we learn ECG representation in alignment with the frozen chest X-ray embeddings. We find in our experiments that un-freezing $f_\text{cxr}$ leads to poor downstream performance. For cross-modal alignment, we use modality specific projection heads $g_\text{cxr}$ and $g_\text{ecg}$ as shown in Fig.~\ref{fig:main_fig} to align the projected embeddings of the two modalities using our supervised cross-modal alignment objective with adaptive hard negative penalization, $\mathcal{L}_\text{AHNPSupCMA}$ (Eqn.~\ref{eqn:ahnpsupcma_term}). We use the \texttt{AdamW} optimizer with peak LR of $10^{-4}$, batch size of 256, linear warm-up for 10 epochs from LR $10^{-5}$, followed by gradual cosine decay to $10^{-5}$. We set weight decay to $10^{-5}$ and clip gradients to 2.5 for training stability. The learning rate for both projection heads is scaled by $0.1$ times relative to that of $f_\text{ecg}$, as they are simple linear layers. To help mitigate the effect of high class imbalance in the dataset, we use \texttt{WeightedRandomSampler} with replacement to over-sample from the minority class ($y_i=1$) to ensure they constitute $25-30\%$ of each batch. To help the model learn generalizable ECG representation, we use the time-domain ECG augmentations described in \citep{raghu_2022_data}. 

\subsubsection{Fine-tuning}
Finally, we discard the projection heads $g_\text{ecg}$, $g_\text{cxr}$ and fine-tune the ECG encoder $f_\text{ecg}$ on the disease labels with a Cross-Entropy objective, and continue to over-sample from the minority class as describe above. We use the \texttt{Adam} optimizer with peak LR of $10^{-5}$, batch size of 256, linear warm-up for 10 epochs from LR $10^{-6}$, followed by gradual cosine decay to $10^{-6}$. We obtain the optimal hyperparameters from the best fine-tuning AUROC results as follows: $\tau=0.01$, \texttt{topk} AHNP, with $k = 7.5\%$, $\alpha = 4.5$ and $\beta=2$, refer Eqn.~\ref{eqn:ahnpsupcma_term}.

We summarize all model and training hyperparameters in Table~\ref{tab:hyperparams}, and the training procedure in Algorithm~\ref{alg:training}.

\section{Experiment and Results}

\begin{table}[h]
    \caption{Performance comparison of \modelname against baselines, along with ablation studies for three pathologies. We also show the metrics for CXR-based classification using $f_\text{cxr}$ at the bottom for reference.}
    \label{tab:results}
    \centering
    \resizebox{\textwidth}{!}{
    \begin{tabular}{lcccccc}
        \toprule
        \textbf{Model} 
        & \multicolumn{2}{c}{\textbf{Cardiomegaly}} 
        & \multicolumn{2}{c}{\textbf{Edema}} 
        & \multicolumn{2}{c}{\textbf{Pleural Effusion}} \\
        & AUROC $\uparrow$ & F1 $\uparrow$
        & AUROC $\uparrow$ & F1 $\uparrow$
        & AUROC $\uparrow$ & F1 $\uparrow$\\
        \midrule
        {Direct Classif. (ECG)} 
        & $72.84 \pm 0.51$ & $30.31 \pm 0.21$
        & $74.89 \pm 0.36$ & $18.11 \pm 0.28$
        & $72.20 \pm 0.38$ & $20.18 \pm 0.19$ \\
        \midrule
        {ResNet-1D}
        & $70.65 \pm 0.22$ & $28.24 \pm 0.41$
        & $68.42 \pm 0.30$ & $17.42 \pm 0.30$
        & $63.51 \pm 0.41$ & $17.99 \pm 0.34$ \\
        {ECCL} \citep{ding_crossmodality_2024}
        & $70.85 \pm 0.14$ & $27.47 \pm 0.13$
        & $67.87 \pm 0.91$ & $13.68 \pm 0.31$
        & $64.12 \pm 0.67$ & $17.81 \pm 0.28$ \\
        {BIOT} \citep{yang_biot_2023}
        & $72.63 \pm 0.02$ & $30.17 \pm 0.05$
        & $69.45 \pm 0.33$ & $13.84 \pm 0.61$
        & $60.23 \pm 0.10$ & $16.52 \pm 0.13$ \\
        {SupCon} \citep{khosla_supervised_2020}
        & $74.46 \pm 0.16$ & $31.42 \pm 0.31$
        & $77.64 \pm 0.38$ & ${20.15 \pm 0.12}$
        & $70.81 \pm 0.20$ & ${25.98 \pm 0.59}$ \\
        {Cross-modal AE} \citep{radhakrishnan_crossmodal_2023}
        & $72.15 \pm 0.13$ & $28.31 \pm 0.79$
        & $68.26 \pm 0.42$ & $13.46 \pm 0.38$
        & $66.08 \pm 0.84$ & $19.72 \pm 0.48$ \\
        \midrule
        {\modelname}\textsubscript{(no SSL, no AHNP)} 
        & $74.66 \pm 0.10$ & $31.19 \pm 0.28$
        & $76.23 \pm 0.19$ & $19.40 \pm 0.08$
        & $72.18 \pm 0.09$ & $22.01 \pm 0.20$ \\
        {\modelname}\textsubscript{(no AHNP)}
        & $74.73 \pm 0.11$ & $31.72 \pm 0.14$
        & $76.78 \pm 0.10$ & $18.76 \pm 0.12$
        & $71.78 \pm 0.23$ & $21.28 \pm 0.34$ \\
        {\modelname}\textsubscript{(no SSL)}
        & $73.08 \pm 0.01$ & $29.67 \pm 0.10$
        & $77.98 \pm 0.30$ & $17.13 \pm 0.11$
        & $72.09 \pm 0.31$ & $22.82 \pm 0.42$ \\
        \midrule
        \textbf{\textcolor{black}{{\modelname} (ours)}}
        & $\mathbf{75.23 \pm 0.04}$ & $\mathbf{31.90 \pm 0.21}$
        & $\mathbf{78.31 \pm 0.25}$ & $\mathbf{{22.39 \pm 0.83}}$
        & $\mathbf{75.51 \pm 0.32}$ & $\mathbf{{29.17 \pm 0.41}}$ \\
        \bottomrule
        \toprule
        {Direct Classif. (CXR)} 
        & $86.27 \pm 0.18$ & $45.70 \pm 0.14$
        & $93.62 \pm 0.13$ & $47.45 \pm 0.21$
        & $93.59 \pm 0.23$ & $61.25 \pm 0.19$ \\
        \bottomrule
    \end{tabular}    
    }
\end{table}

\begin{table}[h]
    \caption{Comparison of adaptive hard negative penalization strategies (\texttt{topk}, \texttt{linear}, \texttt{exp}) across three pathologies.}
    \label{tab:ahnps_combined}
    \centering
    \begin{tabular}{lccc}
        \toprule
        \textbf{Pathology} & \textbf{\texttt{topk}} & \textbf{\texttt{linear}} & \textbf{\texttt{exp}} \\
        \midrule
        \multicolumn{4}{c}{\textbf{AUROC}} \\
        \midrule
        Cardiomegaly & $\mathbf{75.23 \pm 0.04}$ & $73.92 \pm 0.12$ & $75.02 \pm 0.08$ \\
        Edema        & ${77.55 \pm 0.23}$ & $77.89 \pm 0.18$ & $\mathbf{78.31 \pm 0.25}$ \\
        Pleural Effusion & $\mathbf{75.51 \pm 0.32}$ & $74.29 \pm 0.11$ & $74.40 \pm 0.32$ \\
        \midrule
        \multicolumn{4}{c}{\textbf{F1}} \\
        \midrule
        Cardiomegaly & $\mathbf{31.90 \pm 0.21}$ & $30.28 \pm 0.16$ & $30.92 \pm 0.18$ \\
        Edema        & $20.19 \pm 0.62$ & $21.34 \pm 0.18$ & $\mathbf{22.39 \pm 0.83}$ \\
        Pleural Effusion & $\mathbf{29.17 \pm 0.41}$ & $25.69 \pm 0.48$ & $27.54 \pm 0.28$ \\
        \bottomrule
    \end{tabular}
\end{table}

\begin{figure*}
    \centering
    \includegraphics[scale=0.9]{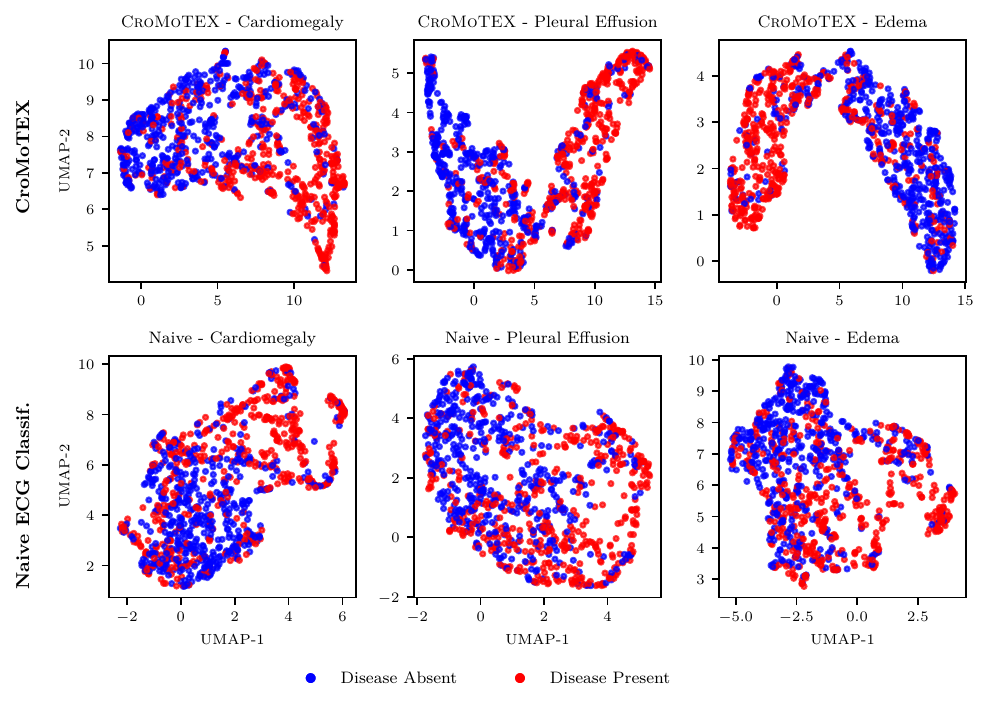}
    \caption{UMAP \citep{mcinnes2018umap} projections of ECG embeddings from our proposed model (\modelname) versus a naive ECG classification baseline. The embeddings produced by \modelname exhibit improved class separation, reflecting more semantically structured representations. This aligns with the superior classification performance reported in Table~\ref{tab:results}.}
    \label{fig:umap}
\end{figure*}

\begin{table}[h]
    \caption{Summary of hyper-parameters used for the proposed model.}
    \label{tab:hyperparams}
    \centering
    \begin{tabular}{llll}
        \toprule
        \multicolumn{2}{c}{\textbf{ECG Encoder (Architecture)}} & \multicolumn{2}{c}{\textbf{Cross-modal Alignment}} \\
        \midrule
        Input Shape & [12, 1000] & Batch Size & 256 \\
        Kernel Size & 5 & LR (Peak) & $10^{-4}$ \\
        Stride1 & 5 & LR (Start) & $10^{-5}$ \\
        Stride2 & 1 & Sched. & Cos Ann. w/ LR \\
        Intmdt. Dim & 128 & Temp. $\tau$ & 0.01 \\
        Embed Dim & 256 & AHNP & \texttt{topk, exp} \\
        Num Heads & 8 & AHNP $k$ & 7.5\% \\
        Num Layers & 4 & AHNP $\alpha$ & 4.5 \\
        Activ. (Conv1D) & ReLU & Pos. Pair Wt. $\beta$ & 3 \\
        Activ. (Transf.) & GELU & Wt. Decay & $10^{-5}$ \\
        Classif. Dim. & 256 & \multicolumn{2}{c}{\textbf{Fine-tuning}} \\
        Classif. Layers & 4 & Batch Size & 256 \\
        Classif. Dropout & 0.1 & LR (Peak) & $10^{-5}$ \\
        & & LR (Start) & $10^{-6}$ \\
        & & Sched. & Cos Ann. w/ LR \\
        & & Loss & Cross-Entropy \\
        & & Optimizer & Adam \\
        \bottomrule
    \end{tabular}
\end{table}

\subsection{Datasets}
\label{subsec:datasets}
We perform experiments using the MIMIC-IV family of datasets \citep{johnson_2023_mimic,goldberger_2000_physiobank} consisting of: MIMIC-IV-ECG, a dataset of approximately $800,000$ diagnostic electrocardiograms; MIMIC-CXR \citep{johnson_2019_mimic}, a dataset of $377,110$ chest X-rays with associated radiology reports, and disease labels derived from CheXpert labeler, which serves as the imaging dataset in our experiments; and MIMIC-IV which serves as the core patient database helping us link chest X-rays to the corresponding ECGs. We merge the three datasets to obtain a single table $\mathcal{T}_\text{full}: \left\{\left(e_i, x_i, y_i, \text{\texttt{subject{\textunderscore}id}}_i\right)\right\}_{i=1}^N$, where \texttt{subject{\textunderscore}id} uniquely identifies a patient. Note that $\mathcal{T}_\text{full}$ contains rows with missing ECGs $e_i$, as not every patient who had a chest X-ray also had an ECG during the same hospital visit. We then perform a random train-validation-test split of $70:10:20$ on unique \texttt{subject{\textunderscore}id}s to obtain the training ($\mathcal{T}_\text{full}^\text{train}$), validation ($\mathcal{T}_\text{full}^\text{val}$) and testing ($\mathcal{T}_\text{full}^\text{test}$) data splits. For performing cross-modal alignment, we obtain paired subsets of the training and validation data by dropping rows with missing ECGs: $\mathcal{T}_\text{paired}^\text{split} = \{\left(e,x,y,\text{\texttt{subject{\textunderscore}id}}\right): e \ne \text{\texttt{NULL}}, \left(e,x,y,\text{\texttt{subject{\textunderscore}id}}\right) \in \mathcal{T}_\text{full}^{\text{split}}\}$ for $\text{split} \in \left\{\text{train}, \text{val}\right\}$. For self-supervised pre-training of the ECG encoder, we create training and validation datasets $\mathcal{T}_\text{ecg}^\text{train}$ and $\mathcal{T}_\text{ecg}^\text{val}$ consisting of all ECGs in MIMIC-IV-ECG, except those belonging to \texttt{subject{\textunderscore}id}s present in $\mathcal{T}_\text{full}^\text{test}$.

We pre-process the raw ECGs by first down-sampling to $100\thinspace{\text{Hz}}$. We then remove baseline-wander \citep{lenis_2017_comparison,thapa_more_2024} and normalize each ECG lead to the range $\left[-1, 1\right]$ to obtain an array of shape \texttt{[12, 1000]}.
For the self-supervised pre-training of ECG encoder (Fig.~\ref{fig:main_fig}), we utilize the augmentations based on vector-cardiogram (VCG) \citep{gopal_3kg_2021}. A VCG is obtained from an ECG through the inverse Dower's transform $D^{-1}$ \citep{gopal_3kg_2021}, and represents the electrical activity of the heart in 3D spatial coordinates. We obtain an augmented ECG as $\operatorname{V}(e_i) = DSRD^{-1}\left(e_i\right)$ where D and $D^{-1}$ represent the Dower's and inverse Dower's transform respectively, $S \in \mathbb{R}^{3\times3}$ is a random scaling matrix and $R \in \mathbb{R}^{3\times3}$ is a random rotation matrix. We then apply random time masking on the resulting ECG. In the cross-modal alignment stage, we use the time-domain ECG augmentations proposed in \citep{raghu_2022_data}.
We convert the chest X-ray images to gray scale, re-normalize pixel values to the range $\left[-1024, 1024\right]$, perform center crop, and resize them to $224\times 224$ pixels.

\subsection{Results: Cross-Modal Transfer for ECG-Based Diagnosis}
As shown in Fig.~\ref{fig:main_fig}, we perform pre-training, cross-modal alignment, and fine-tuning to learn ECG representations using chest X-ray-derived knowledge. We evaluate \modelname on the downstream classification task of detecting three different pathologies using ECG alone: Cardiomegaly, Edema and Pleural Effusion. We use the following metrics to evaluate classification performance: AUROC (Area Under Receiver Operating Characteristics curve)  and F1 score (harmonic mean of precision and recall).

\textbf{Baselines and Ablations}: Given the limited prior research on cross-modal transfer between chest X-rays and ECGs, we evaluate our approach by comparing \modelname against the following baselines: \textbf{SupCon} \citep{khosla_supervised_2020}, a contrastive learning framework that adds explicit supervision from labels to enhance task-relevant representation learning; \textbf{ECCL} \citep{ding_crossmodality_2024} and \textbf{Cross-modal AE} \citep{radhakrishnan_crossmodal_2023} which are cross-modal transfer frameworks from Cardiac Magnetic Resonance (CMR) images to ECGs; \textbf{BIOT} \citep{yang_biot_2023}, a Transformer-based architecture for bio-signals such as ECGs; and \textbf{ResNet-1D}, a popular architecture choice for encoding ECGs \citep{liu_zeroshot_2024}. In addition to these comparisons, we conduct ablation studies to systematically examine the contributions of key components in our framework. 

As seen in Table~\ref{tab:results}, \modelname outperforms the baseline models in the classification of all three pathologies, under both AUROC and F1 score metrics. These results show that our approach significantly improves the detection of these pathologies over direct ECG-based classification, as a result of the cross-modal alignment with CXR. The use of SupCon loss, instead of our AHNP loss with the same architecture yields close, second best results, showing the benefits of supervised alignment in downstream classification performance. On the other hand ECCL and Cross-modal AE, which do not use supervision from the labels, do not deliver good classification performance. Although ECCL combines contrastive loss and classification loss during alignment, this leads to poorer results than using explicit supervision in the contrastive objective. BIOT (Biosignal Transformer) and ResNet-1D are other popular choices for encoding ECGs. As both BIOT and \modelname use a Transformer as part of ECG encoder backbone, BIOT's under-performance indicates the efficacy of our patch embedding convolution layers described in Sec.~\ref{subsec:model_and_training_details}. While purely convolution-based architectures such as ResNet-1D have been successfully used for encoding ECGs, we see that the Transformer-based \modelname performs significantly better across the three pathologies, leading us to believe that the attention mechanism helps achieve superior encoding of spatiotemporal relations in an ECG.

In addition to quantitative metrics, we visualize the structure of the learned ECG embeddings using UMAP \citep{mcinnes2018umap} projections in Fig.~\ref{fig:umap}. For each pathology, we compare embeddings from a naive ECG classifier with those produced by \modelname after cross-modal alignment with chest X-rays and subsequent fine-tuning as described in Sec.~\ref{subsec:model_and_training_details}. The projections reveal that \modelname yields significantly improved class separation in the latent space, with visibly more distinct clustering of positive and negative samples. In contrast, the naive classifier produces relatively poorly structured embeddings. These observations provide qualitative support for the effectiveness of our contrastive alignment strategy, and further corroborate the improvements seen in classification performance.

To further understand the impact of different adaptive hard negative penalization (AHNP) strategies, we compare the performance of \texttt{topk}, \texttt{linear}, and \texttt{exp} weighting schemes across three pathologies in Table~\ref{tab:ahnps_combined}. We observe that the \texttt{topk} strategy yields the best AUROC and F1 scores for Cardiomegaly and Pleural Effusion, indicating its effectiveness in these relatively well-separated tasks. Interestingly, the \texttt{exp} strategy outperforms the others for Edema, suggesting that its emphasis on progressively harder negatives may be better suited for subtler or less separable conditions. While the optimal AHNP strategy may vary by pathology, we find that using \texttt{topk} consistently across all tasks still achieves performance highly competitive with the best baseline per pathology. These results demonstrate the flexibility of our framework to accommodate multiple modes of hard negative penalization and highlight \texttt{topk} as a strong default choice.

Finally, the ablations (Table~\ref{tab:results}) clearly outline the key contributions of various components of our framework and training strategy. We see that both AHNP and the Self-Supervised Learning (SSL pre-training, described in Sec.~\ref{subsec:pretraining}) of ECG encoder play a crucial role in improving the performance of \modelname. While neither component alone helps us outperform the baselines, SSL pre-training combined with cross-modal alignment using AHNP loss help us achieve state-of-the-art performance across all three diseases. 

\section{Discussion}
Our findings have meaningful clinical implications, particularly in the context of expanding access to early diagnostics. While chest X-rays remain the gold standard for diagnosing conditions like cardiomegaly, pleural effusion, and edema, their limited availability in non-hospital settings poses a barrier to timely detection. ECGs, on the other hand, are more readily acquired--nowadays via wearable smart devices--making them attractive for large-scale, accessible screening. By transferring diagnostic knowledge from CXRs to ECGs, \modelname enables improved ECG-based detection of these conditions, potentially supporting earlier intervention in resource-constrained or ambulatory care scenarios.

In evaluating \modelname across multiple cardiopulmonary pathologies, we observe consistent improvements over strong baselines such as ECCL, Cross-Modal Autoencoders, and BIOT. The combination of self-supervised ECG pretraining with our novel AHNP-based supervised contrastive loss enables more task-relevant and robust ECG representations. These improvements are not only quantitative but also evident in qualitative embedding separability, as seen in our UMAP visualizations.

\section{Conclusion}
We introduced \modelname, a supervised cross-modal learning framework that transfers diagnostic knowledge from chest X-rays to ECGs to improve ECG-based disease detection. By incorporating label supervision and our proposed AHNP loss, \modelname achieves state-of-the-art performance on multiple cardiopulmonary pathologies and outperforms existing baselines. Our ablation studies confirm the complementary roles of self-supervised ECG pretraining and adaptive contrastive alignment in achieving these gains. 

However, despite these advances, we observe a significant performance gap between our cross-modal transfer model and direct CXR-based classification, indicating room for further improvements in ECG-CXR alignment. Bridging this gap through enhanced alignment strategies, better modality fusion techniques, or larger-scale multimodal pretraining remains a promising direction for future research.

\bibliography{refs}

\begin{thebibliography}{30}
\providecommand{\natexlab}[1]{#1}
\providecommand{\url}[1]{\texttt{#1}}
\expandafter\ifx\csname urlstyle\endcsname\relax
  \providecommand{\doi}[1]{doi: #1}\else
  \providecommand{\doi}{doi: \begingroup \urlstyle{rm}\Url}\fi

\bibitem[Chen et~al.(2020)Chen, Kornblith, Norouzi, and
  Hinton]{chen_simple_2020}
T.~Chen, S.~Kornblith, M.~Norouzi, and G.~Hinton.
\newblock A {{Simple Framework}} for {{Contrastive Learning}} of {{Visual
  Representations}}.
\newblock In \emph{Proceedings of the 37th {{International Conference}} on
  {{Machine Learning}}}, pages 1597--1607. PMLR, Nov. 2020.

\bibitem[Cheng et~al.(2020)Cheng, Goh, Dogrusoz, Tuzel, and
  Azemi]{cheng_2020_subject}
J.~Y. Cheng, H.~Goh, K.~Dogrusoz, O.~Tuzel, and E.~Azemi.
\newblock Subject-aware contrastive learning for biosignals.
\newblock \emph{arXiv preprint arXiv:2007.04871}, 2020.

\bibitem[Cohen et~al.(2022)Cohen, Viviano, Bertin, Morrison, Torabian,
  Guarrera, Lungren, Chaudhari, Brooks, Hashir,
  et~al.]{cohen_2022_torchxrayvision}
J.~P. Cohen, J.~D. Viviano, P.~Bertin, P.~Morrison, P.~Torabian, M.~Guarrera,
  M.~P. Lungren, A.~Chaudhari, R.~Brooks, M.~Hashir, et~al.
\newblock Torchxrayvision: A library of chest x-ray datasets and models.
\newblock In \emph{International Conference on Medical Imaging with Deep
  Learning}, pages 231--249. PMLR, 2022.

\bibitem[Ding et~al.(2024)Ding, Hu, Li, Zhang, Wu, Xiang, Li, Liu, Chu, and
  Huang]{ding_crossmodality_2024}
Z.~Ding, Y.~Hu, Z.~Li, H.~Zhang, F.~Wu, Y.~Xiang, T.~Li, Z.~Liu, X.~Chu, and
  Z.~Huang.
\newblock Cross-{{Modality Cardiac Insight Transfer}}: {{A Contrastive Learning
  Approach}} to~{{Enrich ECG}} with~{{CMR Features}}.
\newblock In \emph{Medical {{Image Computing}} and {{Computer Assisted
  Intervention}} -- {{MICCAI}} 2024}, pages 109--119. Springer Nature
  Switzerland, 2024.

\bibitem[Goldberger et~al.(2000)Goldberger, Amaral, Glass, Hausdorff, Ivanov,
  Mark, Mietus, Moody, Peng, and Stanley]{goldberger_2000_physiobank}
A.~L. Goldberger, L.~A. Amaral, L.~Glass, J.~M. Hausdorff, P.~C. Ivanov, R.~G.
  Mark, J.~E. Mietus, G.~B. Moody, C.-K. Peng, and H.~E. Stanley.
\newblock Physiobank, physiotoolkit, and physionet: components of a new
  research resource for complex physiologic signals.
\newblock \emph{Circulation}, 101\penalty0 (23):\penalty0 e215--e220, 2000.

\bibitem[Gopal et~al.(2021)Gopal, Han, Raghupathi, Ng, Tison, and
  Rajpurkar]{gopal_3kg_2021}
B.~Gopal, R.~Han, G.~Raghupathi, A.~Ng, G.~Tison, and P.~Rajpurkar.
\newblock {{3KG}}: {{Contrastive Learning}} of 12-{{Lead Electrocardiograms}}
  using {{Physiologically-Inspired Augmentations}}.
\newblock In \emph{Proceedings of {{Machine Learning}} for {{Health}}}, pages
  156--167. PMLR, Nov. 2021.

\bibitem[Grill et~al.(2020)Grill, Strub, Altch{\'e}, Tallec, Richemond,
  Buchatskaya, Doersch, Avila~Pires, Guo, Gheshlaghi~Azar,
  et~al.]{grill_2020_bootstrap}
J.-B. Grill, F.~Strub, F.~Altch{\'e}, C.~Tallec, P.~Richemond, E.~Buchatskaya,
  C.~Doersch, B.~Avila~Pires, Z.~Guo, M.~Gheshlaghi~Azar, et~al.
\newblock Bootstrap your own latent-a new approach to self-supervised learning.
\newblock \emph{Advances in neural information processing systems},
  33:\penalty0 21271--21284, 2020.

\bibitem[Hager et~al.(2023)Hager, Menten, and Rueckert]{hager_best_2023}
P.~Hager, M.~J. Menten, and D.~Rueckert.
\newblock Best of {{Both Worlds}}: {{Multimodal Contrastive Learning With
  Tabular}} and {{Imaging Data}}.
\newblock \emph{Proceedings of the IEEE/CVF Conference on Computer Vision and
  Pattern Recognition}, pages 23924--23935, 2023.

\bibitem[Hayat et~al.(2023)Hayat, Geras, and Shamout]{hayat_medfuse_2023}
N.~Hayat, K.~J. Geras, and F.~E. Shamout.
\newblock {{MedFuse}}: {{Multi-modal}} fusion with clinical time-series data
  and chest {{X-ray}} images, Mar. 2023.

\bibitem[He et~al.(2020)He, Fan, Wu, Xie, and Girshick]{he_2020_momentum}
K.~He, H.~Fan, Y.~Wu, S.~Xie, and R.~Girshick.
\newblock Momentum contrast for unsupervised visual representation learning.
\newblock In \emph{Proceedings of the IEEE/CVF conference on computer vision
  and pattern recognition}, pages 9729--9738, 2020.

\bibitem[Huang et~al.(2021)Huang, Shen, Lungren, and Yeung]{huang_gloria_2021}
S.-C. Huang, L.~Shen, M.~P. Lungren, and S.~Yeung.
\newblock {{GLoRIA}}: {{A Multimodal Global-Local Representation Learning
  Framework}} for {{Label-Efficient Medical Image Recognition}}.
\newblock \emph{Proceedings of the IEEE/CVF International Conference on
  Computer Vision}, pages 3942--3951, 2021.

\bibitem[Jia et~al.(2021)Jia, Yang, Xia, Chen, Parekh, Pham, Le, Sung, Li, and
  Duerig]{jia_2021_scaling}
C.~Jia, Y.~Yang, Y.~Xia, Y.-T. Chen, Z.~Parekh, H.~Pham, Q.~Le, Y.-H. Sung,
  Z.~Li, and T.~Duerig.
\newblock Scaling up visual and vision-language representation learning with
  noisy text supervision.
\newblock In \emph{International conference on machine learning}, pages
  4904--4916. PMLR, 2021.

\bibitem[Johnson et~al.(2019)Johnson, Pollard, Berkowitz, Greenbaum, Lungren,
  Deng, Mark, and Horng]{johnson_2019_mimic}
A.~E. Johnson, T.~J. Pollard, S.~J. Berkowitz, N.~R. Greenbaum, M.~P. Lungren,
  C.-y. Deng, R.~G. Mark, and S.~Horng.
\newblock Mimic-cxr, a de-identified publicly available database of chest
  radiographs with free-text reports.
\newblock \emph{Scientific data}, 6\penalty0 (1):\penalty0 317, 2019.

\bibitem[Johnson et~al.(2023)Johnson, Bulgarelli, Shen, Gayles, Shammout,
  Horng, Pollard, Hao, Moody, Gow, et~al.]{johnson_2023_mimic}
A.~E. Johnson, L.~Bulgarelli, L.~Shen, A.~Gayles, A.~Shammout, S.~Horng, T.~J.
  Pollard, S.~Hao, B.~Moody, B.~Gow, et~al.
\newblock Mimic-iv, a freely accessible electronic health record dataset.
\newblock \emph{Scientific data}, 10\penalty0 (1):\penalty0 1, 2023.

\bibitem[Khosla et~al.(2020)Khosla, Teterwak, Wang, Sarna, Tian, Isola,
  Maschinot, Liu, and Krishnan]{khosla_supervised_2020}
P.~Khosla, P.~Teterwak, C.~Wang, A.~Sarna, Y.~Tian, P.~Isola, A.~Maschinot,
  C.~Liu, and D.~Krishnan.
\newblock Supervised {{Contrastive Learning}}.
\newblock In \emph{Advances in {{Neural Information Processing Systems}}},
  volume~33, pages 18661--18673. Curran Associates, Inc., 2020.

\bibitem[King et~al.(2023)King, Yang, and Mortazavi]{king_multimodal_2023}
R.~King, T.~Yang, and B.~J. Mortazavi.
\newblock Multimodal {{Pretraining}} of {{Medical Time Series}} and {{Notes}}.
\newblock In \emph{Proceedings of the 3rd {{Machine Learning}} for {{Health
  Symposium}}}, pages 244--255. PMLR, Dec. 2023.

\bibitem[Kiyasseh et~al.(2021)Kiyasseh, Zhu, and Clifton]{kiyasseh_2021_clocs}
D.~Kiyasseh, T.~Zhu, and D.~A. Clifton.
\newblock Clocs: Contrastive learning of cardiac signals across space, time,
  and patients.
\newblock In \emph{International Conference on Machine Learning}, pages
  5606--5615. PMLR, 2021.

\bibitem[Lenis et~al.(2017)Lenis, Pilia, Loewe, Schulze, and
  D{\"o}ssel]{lenis_2017_comparison}
G.~Lenis, N.~Pilia, A.~Loewe, W.~H. Schulze, and O.~D{\"o}ssel.
\newblock Comparison of baseline wander removal techniques considering the
  preservation of st changes in the ischemic ecg: a simulation study.
\newblock \emph{Computational and mathematical methods in medicine},
  2017\penalty0 (1):\penalty0 9295029, 2017.

\bibitem[Liu et~al.(2024)Liu, Wan, Ouyang, Shah, Bai, and
  Arcucci]{liu_zeroshot_2024}
C.~Liu, Z.~Wan, C.~Ouyang, A.~Shah, W.~Bai, and R.~Arcucci.
\newblock Zero-{{Shot ECG Classification}} with {{Multimodal Learning}} and
  {{Test-time Clinical Knowledge Enhancement}}.
\newblock In \emph{Forty-First {{International Conference}} on {{Machine
  Learning}}}, June 2024.

\bibitem[Liu et~al.(2019)Liu, Zhao, Fei, Zhang, Wang, and Yu]{liu_2019_align}
J.~Liu, G.~Zhao, Y.~Fei, M.~Zhang, Y.~Wang, and Y.~Yu.
\newblock Align, attend and locate: Chest x-ray diagnosis via contrast induced
  attention network with limited supervision.
\newblock In \emph{Proceedings of the IEEE/CVF International Conference on
  Computer Vision}, pages 10632--10641, 2019.

\bibitem[McInnes et~al.(2018)McInnes, Healy, and Melville]{mcinnes2018umap}
L.~McInnes, J.~Healy, and J.~Melville.
\newblock Umap: Uniform manifold approximation and projection for dimension
  reduction.
\newblock \emph{arXiv preprint arXiv:1802.03426}, 2018.

\bibitem[Oord et~al.(2018)Oord, Li, and Vinyals]{oord_2018_representation}
A.~v.~d. Oord, Y.~Li, and O.~Vinyals.
\newblock Representation learning with contrastive predictive coding.
\newblock \emph{arXiv preprint arXiv:1807.03748}, 2018.

\bibitem[Radford et~al.(2021)Radford, Kim, Hallacy, Ramesh, Goh, Agarwal,
  Sastry, Askell, Mishkin, Clark, Krueger, and
  Sutskever]{radford_learning_2021}
A.~Radford, J.~W. Kim, C.~Hallacy, A.~Ramesh, G.~Goh, S.~Agarwal, G.~Sastry,
  A.~Askell, P.~Mishkin, J.~Clark, G.~Krueger, and I.~Sutskever.
\newblock Learning {{Transferable Visual Models From Natural Language
  Supervision}}.
\newblock In \emph{Proceedings of the 38th {{International Conference}} on
  {{Machine Learning}}}, pages 8748--8763. PMLR, July 2021.

\bibitem[Radhakrishnan et~al.(2023)Radhakrishnan, Friedman, Khurshid, Ng,
  Batra, Lubitz, Philippakis, and Uhler]{radhakrishnan_crossmodal_2023}
A.~Radhakrishnan, S.~F. Friedman, S.~Khurshid, K.~Ng, P.~Batra, S.~A. Lubitz,
  A.~A. Philippakis, and C.~Uhler.
\newblock Cross-modal autoencoder framework learns holistic representations of
  cardiovascular state.
\newblock \emph{Nature Communications}, 14\penalty0 (1):\penalty0 2436, Apr.
  2023.
\newblock ISSN 2041-1723.
\newblock \doi{10.1038/s41467-023-38125-0}.
\newblock URL \url{https://www.nature.com/articles/s41467-023-38125-0}.

\bibitem[Raghu et~al.(2022)Raghu, Shanmugam, Pomerantsev, Guttag, and
  Stultz]{raghu_2022_data}
A.~Raghu, D.~Shanmugam, E.~Pomerantsev, J.~Guttag, and C.~M. Stultz.
\newblock Data augmentation for electrocardiograms.
\newblock In \emph{conference on health, inference, and learning}, pages
  282--310. PMLR, 2022.

\bibitem[Sowrirajan et~al.(2021)Sowrirajan, Yang, Ng, and
  Rajpurkar]{sowrirajan_moco_2021}
H.~Sowrirajan, J.~Yang, A.~Y. Ng, and P.~Rajpurkar.
\newblock {{MoCo Pretraining Improves Representation}} and {{Transferability}}
  of {{Chest X-ray Models}}.
\newblock In \emph{Proceedings of the {{Fourth Conference}} on {{Medical
  Imaging}} with {{Deep Learning}}}, pages 728--744. PMLR, Aug. 2021.

\bibitem[Thapa et~al.(2024)Thapa, Howlader, Bhattacharjee, and
  {le}]{thapa_more_2024}
S.~Thapa, K.~Howlader, S.~Bhattacharjee, and W.~{le}.
\newblock {{MoRE}}: {{Multi-Modal Contrastive Pre-training}} with
  {{Transformers}} on {{X-Rays}}, {{ECGs}}, and {{Diagnostic Report}}, Oct.
  2024.

\bibitem[Vaswani et~al.(2017)Vaswani, Shazeer, Parmar, Uszkoreit, Jones, Gomez,
  Kaiser, and Polosukhin]{vaswani_2017_attention}
A.~Vaswani, N.~Shazeer, N.~Parmar, J.~Uszkoreit, L.~Jones, A.~N. Gomez,
  {\L}.~Kaiser, and I.~Polosukhin.
\newblock Attention is all you need.
\newblock \emph{Advances in neural information processing systems}, 30, 2017.

\bibitem[Yang et~al.(2023)Yang, Westover, and Sun]{yang_biot_2023}
C.~Yang, M.~B. Westover, and J.~Sun.
\newblock {{BIOT}}: {{Biosignal Transformer}} for {{Cross-data Learning}} in
  the {{Wild}}.
\newblock \emph{37th Conference on Neural Information Processing Systems
  (NeurIPS 2023)}, 2023.

\bibitem[Yao et~al.(2024)Yao, Yin, Cheung, Liu, and Qin]{yao_drfuse_2024}
W.~Yao, K.~Yin, W.~K. Cheung, J.~Liu, and J.~Qin.
\newblock {{DrFuse}}: {{Learning Disentangled Representation}} for {{Clinical
  Multi-Modal Fusion}} with {{Missing Modality}} and {{Modal Inconsistency}}.
\newblock \emph{Proceedings of the AAAI Conference on Artificial Intelligence},
  38\penalty0 (15):\penalty0 16416--16424, Mar. 2024.
\newblock ISSN 2374-3468.

\end{thebibliography}

\end{document}